\documentclass[12pt]{article}
\usepackage[natbibapa]{apacite}
\usepackage{amsmath}
\usepackage{times}
\usepackage{graphicx}
\usepackage{color}
\usepackage{multirow}
\usepackage{rotating}
\usepackage{bbm}
\usepackage{latexsym}
\usepackage{hyperref}
\usepackage{bm}

\newcommand{\captionfonts}{\normalsize}

\makeatletter  
\long\def\@makecaption#1#2{%
  \vskip\abovecaptionskip
  \sbox\@tempboxa{{\captionfonts #1: #2}}%
  \ifdim \wd\@tempboxa >\hsize
    {\captionfonts #1: #2\par}
  \else
    \hbox to\hsize{\hfil\box\@tempboxa\hfil}%
  \fi
  \vskip\belowcaptionskip}
\makeatother   

\usepackage{enumitem}

\usepackage{graphicx}
\usepackage{epstopdf}
\usepackage{subfig}

\usepackage{amsmath}
\usepackage{tikz}
\usepackage{mathdots}
\usepackage{yhmath}
\usepackage{cancel}
\usepackage{color}
\usepackage{siunitx}
\usepackage{array}
\usepackage{multirow}
\usepackage{amssymb}
\usepackage{gensymb}
\usepackage{tabularx}
\usepackage{booktabs}
\usetikzlibrary{fadings}

\usepackage[]{soul} % highlighting text
\soulregister\citep7
\soulregister\citet7
\soulregister\ref7

\renewcommand{\thefootnote}{\normalsize \fnsymbol{footnote}} 

\begin{document}

% \listoftodos[Notes]

\newpage

\hspace{13.9cm}1

\ \vspace{20mm}\\

{\LARGE The emergence of visual semantics through communication games}
% %Unsupervised Feature Learning
% %contrastive predictive coding, .....see summary paper 

\ \\
{\bf \large Daniela Mihai$\footnotemark[1]$, Jonathon Hare$\footnotemark[1]$}\\
{$\footnotemark[1]$Vision Learning and Control, Electronics and Computer Science, University of Southampton.}\\

\renewcommand{\thefootnote}{\arabic{footnote}} 

%\ \\[-2mm]

\thispagestyle{empty}
\markboth{}{Emergent Communication in Reference Games}
\ \vspace{-0mm}\\
%
%Abstract
\begin{center} {\bf Abstract} \end{center}
The emergence of communication systems between agents which learn to play referential signalling games with realistic images has attracted a lot of attention recently. The majority of work has focused on using fixed, pretrained image feature extraction networks which potentially bias the information the agents learn to communicate. In this work, we consider a signalling game setting in which a `sender' agent must communicate the information about an image to a `receiver' who must select the correct image from many distractors. We investigate the effect of the feature extractor's weights and of the task being solved on the visual semantics learned by the models. We first demonstrate to what extent the use of pretrained feature extraction networks inductively bias the visual semantics conveyed by emergent communication channel and quantify the visual semantics that are induced.

We then go on to explore ways in which inductive biases can be introduced to encourage the emergence of semantically meaningful communication without the need for any form of supervised pretraining of the visual feature extractor. We impose various augmentations to the input images and additional tasks in the game with the aim to induce visual representations which capture conceptual properties of images. Through our experiments, we demonstrate that communication systems which capture visual semantics can be learned in a completely self-supervised manner by playing the right types of game. Our work bridges a gap between emergent communication research and self-supervised feature learning.

\section{Introduction}
% \todo[inline]{Open with a para that succinctly describes what we want to explore - does the linking of emecomm with feature learning allow the emergence of visual semantics in the communication channel?}
Deep-agent emergent language research aims to develop agents that can cooperate with others, including humans. To achieve this goal, these agents necessarily communicate with particular protocols through communication channels. In emergent-communication research, the communication protocols are learned by the agents, and researchers often investigate how these protocols compare to natural human languages.  In this paper, we study the emergence of visual semantics in such learned communication protocols, in the context of referential signalling games~\citep{Lewis1969-LEWCAP-4}. Although previous research has looked into how pre-linguistic conditions, such as the input representation (either symbolic or raw pixel input) \citep{lazaridou2018emergence}, affect the nature of the communication protocol, we highlight some features of the referential game that can improve the semantics, and hence push it towards a more natural form, and away from a pure image-hashing solution that could na\"ively solve the game perfectly. We then explore the effects of linking language learning with feature learning in a completely self-supervised setting where no information on the objects present in a scene is provided to the model at any point. We thus seek to build a bridge between recent research in self-supervised feature learning with recent advances in self-supervised game play with emergent communication channels.

The idea that agents might learn language by playing visually grounded games has a long history~\citep{cangelosi2002simulating, steels2012experiments}.  Research in this space has recently had something of a resurgence with the introduction of a number of models that simulate the play of \textit{referential} games~\citep{Lewis1969-LEWCAP-4} using realistic visual inputs~\citep{lazaridou2017multi,Havrylov2017,lee2017emergent}. On one hand, these works have shown that the agents can learn to successfully communicate to play these games; however, on the other hand, there has been much discussion as to whether the agents are really learning a communication system grounded in what humans would consider to be the semantics of visual scenes. \citet{BouchacourtB18} highlight this issue in the context of a pair of games designed by~\citet{lazaridou2017multi} which involved the sender and receiver agents being presented with pairs of images. They show that the internal representations of the agents are perfectly aligned, which allows them to successfully play the game but does not enforce capturing conceptual properties. Moreover, when the same game is played with images made up of random noise, the agents still succeed at communicating, which suggests that they agree on and rely on incomprehensible low-level properties of the input which drift away from human-interpretable properties. This finding should perhaps not be so surprising; it is clear to see that one easy way for agents to successfully play these visual communication games would be by developing schemes which create hash-codes from the visual content at very low levels (perhaps even at the pixel level).

\citet{Havrylov2017} explored a different, and potentially harder, game than that proposed by~\citet{lazaridou2017multi}. In their game (see Section~\ref{Section:ExperimentalSetup} for full details), the sender sees the target image and the receiver sees a batch of images formed of a number of distractor images plus the target one. The sender agent is then allowed to send a variable-length message, up to a maximum length, from a fixed vocabulary to the receiver. The later then needs to use that message to identify the target. As opposed to \citet{lazaridou2017multi}'s game in which both agents see only a pair of images, this setting requires the message to include information that will allow the receiver to pick the target image from a batch of 128 images. In their work, they show some qualitative examples in which it does appear that the generated language does in some way convey the visual semantics of the scene (in terms of `objectness` --- correlations between the sequences of tokens of the learnt language and objects, as perceived by humans, known to exist within the images). There are however many open questions from this analysis; one of the key questions is to what extent the ImageNet-pretrained VGG-16 CNN \citep{simonyan2014very} used in the model is affecting the language protocol that emerges. 

In this work, we explore visual semantics in the context of \citet{Havrylov2017}'s game by carefully controlling the visual feature extractor that is used and augmenting the game play in different ways. We seek to explore what factors encourage the emergent language to convey visual semantics rather than falling back to a communication system that just learns hashes of the input images. More concretely, we:
\begin{itemize}
    \item Study the effect of different weights in the CNN used to generate the features (pretrained on ImageNet and frozen as in the original work, randomly initialised and frozen, and, learned end-to-end in the model). We find that models with a feature extractor pretrained in a supervised way capture the most semantic content in the emergent protocol.
    \item Investigate the effect of augmentations that make the game harder by changing the image given to the sender (adding noise and/or random rotations), but not the receiver. Overall, adding noise seems to only make the game slightly harder as the communication success drops, while rotation improves the visual semantics metrics.
    \item Explore the effect of independently augmenting the images given to the sender and the receiver (random cropping and resize to the original image size, random rotations and colour distortion), so they do no see the exact same image. We show that it is possible to get a fully learned model that captures similar amounts of semantic notions as a model with a pretrained feature extractor.
    \item Extend the game to include a secondary task (guessing the rotation of the sender's input) in order to assess whether having agents perform more diverse tasks might lead to stronger visual semantics emerging. We find that without a complex sequence of data augmentation transforms and any supervision, a more meaningful communication protocol can emerge between agents that solve multiple tasks.
    \item Analyse the effect of pretraining the feature extractor network in a self-supervised framework before engaging in the multi-task game. We show that solving such a self-supervised task helps ground the emergent protocol without any human supervision and is even more beneficial for the semantic content captured by a fully learned model.
    
\end{itemize}

We draw attention to the fact that other than in the cases where we use pretrained feature extractors, our simulations are completely self-supervised, and there is no explicit signal of what a human would understand as the `visual semantics' given to the models at any point. If our models are to communicate visual semantics through their communication protocols, then they must learn how to extract features that provide suitable information on those semantics from raw image pixel data.

The remainder of this paper is structured as follows: Section~\ref{Section:Related} looks at related work, which necessarily covers a wide range of topics. Section~\ref{Section:ExperimentalSetup} describes our baseline game and model, building upon~\citet{Havrylov2017}. Section~\ref{Section:Experiments} describes and discusses a range of investigations that explore what can make the emergent communication protocol convey more semantically meaningful information. Finally, section~\ref{Section:Concl} concludes by summarising our findings and makes suggestions for ways in which these could be taken further forward in the future.

\section{Related Work}\label{Section:Related}
In this section, we cover the background literature relevant to our work: the emergence of semantic concepts in artificial communication protocols, without previously embedded knowledge from pretrained features. As our work seeks to bridge research in disparate areas, our discussion necessarily crosses a broad range of topics from the `meaning of images' to emergent communication through game play to feature learning, whilst at the same time considering neural architectures that allow us to build models that can be trained. We first discuss the way humans perceive real-world scenes and what it is that one comprehends as visual semantics. We then proceed and give an overview of the history of multi-agent cooperation games which led to the research field of emergent communication. We look at recent advances that allow us to train emergent communication models parameterised by neural networks using gradient-based methods, and end by looking at recent advances in feature learning.

% \todo[inline]{intro para here just describing how we'll present the related work, and give the big picture that we're really interested in looking at what happens if you link emecomm with feature learning}

\subsection{What do humans perceive as `visual semantics'?}\label{Subsec:humanSemantics}
 When presented with a natural image, humans are capable of answering questions about any objects or creatures, and about any relationships between them \citep{biederman2017semantics}. In this work, we focus on the first question, the \emph{what?}, i.e. the object category (or the list of categories). Research on the way humans perceive real-world scenes such as \cite{biederman1972perceiving} talk about the importance of meaningful and coherent context in perceptual recognition of objects. Their study compares the accuracy of identifying a single object in a real-world jumbled scene versus in a coherent scene. On the other hand, theories such as that by \citet{henderson1999high} support the idea the object identification is independent of global scene context. A slightly more recent psychophysical study \citep{fei2007we} shows that humans, in a single glance of a natural image, are capable of recognising and categorising individual objects in the scene and distinguishing between environments, whilst also perceiving more complex features such as activities performed or social interactions. 
 
 Despite the debate between these two and many other models of scene and object perception, it is clear that the notion of `objects' is important in how a scene is understood by a human. Throughout this work we consider an object-based description of natural images (aligned with what humans would consider to be objects or object categories) to be suitable for the measurement of semantics captured by an emergent communication protocol. Our specific measures are detailed in Section~\ref{Subsec:metrics}.

\subsection{Emergent Communication}

\paragraph{Background.} 
The emergence of language in multi-agent settings has traditionally been studied in the language evolution literature which is concerned with the evolution of communication protocols from scratch \citep{steels1997synthetic, nowak1999evolution}. These early works survey mathematical models and software simulations with artificial agents to explore how various aspects of language have begun and continue to evolve. One key finding of \citet{nowak1999evolution} is that signal-object associations are only possible when the information transfer is beneficial for both parties involved, and hence that \emph{cooperation} is a vital prerequisite for language evolution. Our work is inspired by the renewed interest in the field of emergent communication which uses contemporary deep learning methods to train agents on referential communication games \citep{baroni2020rat, chaab2019antieff, li2019ease, lazaridou2018emergence, cao2018emergent, evtimova2017emergent, Havrylov2017, lazaridou2017multi, lee2017emergent, DBLP:journals/corr/MordatchA17, SukhbaatarSF16}. Their works all build toward the long-standing goal of having specialised agents that can interact with each other and with humans to cooperatively solve tasks and hence assist them in the daily life such as going through different chores. 

\paragraph{Protolanguage and Properties.}
Recent work by \citet{baroni2020rat} highlights some of the priorities in current emergent language research and sketches the characteristics of a useful \emph{protolanguage} for deep agents. It draws on the idea from linguistics that human language has gone through several stages before reaching the full-blown form it has today, and it had to start from a limited set of simple constructions \citep{bickerton2014more}. By providing a realistic scenario of a daily interaction between humans and deep agents, \citet{baroni2020rat} emphasises that a useful protolanguage first needs to use words in order to categorise perceptual input; then allow the creation of new words as new concepts are encountered, and only after, deal with predication structures (i.e. between object-denoting words and property-or-action-denoting words). The focus of our work is on the categorisation phase as we explore whether it is possible for deep agents to develop a language which captures visual concepts whilst simultaneously learning features from natural images in a completely self-supervised way.  

In the referential game setting used in our work, the protolanguage is formed of variable-length sequences of discrete tokens, which are chosen from a predefined, fixed vocabulary. The learned protocol is not grounded in any way, such that the messages are not forced to be similar to those of natural language. As described in Section~\ref{Subsec:humanSemantics}, we believe it is a reasonable assumption that if the game were to be played by human agents they would capture the object's category and its properties that help distinguish the target from the distractor images.

\subsection{Games}
Lewis's classic signalling games \citep{Lewis1969-LEWCAP-4} have been extensively studied for language evolution purposes \citep{steels1997synthetic, nowak1999evolution}, but also in game theory under the name `cheap talk' games. These games are coordination problems in which agents must choose one of several alternative actions, but in which, their decisions are influenced by their expectations of other agents' actions. Similar to Lewis's games, image reference games are coordination problems between multiple agents that require a \emph{limited} communication channel through which information can be exchanged for solving a cooperative task. The task usually requires one agent transmitting information about an image, and a second agent guessing the correct image from several others based on the received message \citep{lazaridou2017multi, lazaridou2018emergence, Havrylov2017}. Other examples of cooperative tasks which require communication between multiple agents include: language translation \citep{lee2017emergent}, logic riddles \citep{FoersterAFW16a}, simple dialog \citep{das2017learning} and negotiation \citep{lewis2017deal}.

One of the goals in emergent communication research is for the developed \emph{protolanguage} to receive no, or as little as possible, human supervision. However, reaching coordination between agents solving a cooperative task, while developing a human-friendly communication protocol has been shown to be extremely difficult \citep{lowe2019pitfalls, chaab2019antieff, KotturMLB17}. In these games, the emergent language has no prior meaning (neither semantics nor syntax) and it converges to develop these by learning to solve the task through many trials or attempts. \citet{lee2019countering} proposes a translation task (i.e. encoding a source language sequence and decoding it into a target language) via a third pivot language. They show that auxiliary constraints on this pivot language help to best retain original syntax and semantics. Other approaches \citep{Havrylov2017, lazaridou2017multi, lee2017emergent} directly force the agents to imitate natural language by using pretrained visual vectors, which already encode information about objects. \citet{Lowe*2020On}, on the other hand, discusses the benefits of combining expert knowledge supervision and self-play, with the end goal of making human-in-the-loop language learning algorithms more efficient. 

Our work builds upon \citet{Havrylov2017}'s referential game (which we describe in more detail in Section~\ref{Section:ExperimentalSetup}) but is also trying to learn the feature extractor, in contrast to the original game in which the feature extractor was pretrained on an object classification task. Therefore, the extracted features are not grounded in the natural language. We take inspiration from all the mentioned papers and investigate to which extent the communication protocol can be encouraged to capture semantics and learn a useful feature extractor in a completely self-supervised way by just solving the predetermined task.

\subsection{Differentiable neural models of representation}

The research works in the previous two subsections predominantly utilise models that communicate with sequences of discrete tokens. Particularly in recent work, the token-producing and token-consuming parts of the models are typically modelled with neural architectures such as variants of recurrent neural networks such as LSTMs. One of the biggest challenges with these models is that the production of discrete tokens necessarily involves a sampling step in which the next  token is drawn from a categorical distribution, which is itself parameterised by a neural network. Such a sampling operation is non-differentiable, and thus, until recently, the only way to learn such models was by using reinforcement learning, and in particular unbiased, but high-variance monte-carlo estimation methods such as the REINFORCE algorithm~\citep{Williams1992} and its variants. 

Over the last six years there has been much interest in neural-probabilistic latent variable models, perhaps most epitomised by \citet{kingma2014autoencoding}'s Variational Autoencoder (VAE). The VAE is an autoencoder that models its latent space, not as continuous fixed-length vectors, but as multivariate normal distributions.
% ; that is to say that a single input is mapped by the encoder to a distribution with a mean vector $\bm\mu$ and a (typically diagonal) covariance matrix $\bm\Sigma$, rather than just to a single vector. 
The decoder part of the VAE however only takes a single sample of the distribution as input. Although they contain a discrete stochastic operation in the middle of the network (sampling $\bm y\sim\mathcal{N}(\bm\mu,\bm\Sigma)$), VAEs are able to be trained with gradient descent using what has become popularly known as the \textit{reparameterisation trick} since the publication of the VAE model \citep{kingma2014autoencoding}, although the idea itself is much older~\citep{Williams1992}. 

% Broadly speaking, the reparameterisation trick allows one to compute gradients with respect to the parameters of a distribution being sampled by factoring out those parameters from the stochastic variable (e.g. $\bm y = \bm\mu + \bm\Sigma \bm z$ where $\bm z \sim \mathcal{N}(\bm 0, \bm I)$ in the case of a multivariate normal distribution).

The reparameterisation trick only applies directly when the distribution can be factored into a function that is continuous and differentiable almost everywhere. In 2017 this limitation was addressed independently by two set papers~\citep{DBLP:conf/iclr/MaddisonMT17,DBLP:conf/iclr/JangGP17} that introduced what we now know as the Gumbel Softmax estimator, which is a reparameterisation that allows us to sample a categorical distribution ($t \sim \operatorname{Cat}(p_1, \dots, p_K)\,; \,\sum_i p_i=1$) from its logits $\bm x$. 
% The first step in understanding this estimator is the Gumbel-max trick:
% \begin{equation*}
% t = \underset{ i \in \{1,\cdots,K\} }{\operatorname{argmax}} x_i + z_i    
% \end{equation*}
% where $z_1, \dots z_K$ are i.i.d Gumbel(0,1) variates which can be computed from Uniform variates through $-\log(-\log(-\mathcal{U}(0,1)))$. Clearly $\operatorname{argmax}$ is not differentiable, but it can be replaced with a continuous approximation using the $\operatorname{softargmax}$:
% \begin{equation*}
%   \operatorname{softargmax}(\bm y) = \sum_i \frac{e^{y_i/T}}{\sum_j e^{y_j/T}} i
% \end{equation*}
% where $T$ is the temperature parameter. This relaxation gives us the Gumbel-softmax, a continuous approximation to sampling a categorical distribution. 

One way to utilise this is to use the Gumbel-softmax approximation during training, and replace it with the hard max at test time, however this can often lead to problems because the model can learn to exploit information leaked through the continuous variables during training. A final trick, the straight-through operator, can be used to circumvent this problem~\citep{DBLP:conf/iclr/JangGP17}.
% \begin{equation*}
%   \operatorname{STargmax}(\bm{y}) = \operatorname{softargmax}(\bm{y}) + \operatorname{stopgradient}(\operatorname{argmax}(\bm{y}) - \operatorname{softargmax}(\bm{y})) 
% \end{equation*}
% where $\operatorname{stopgradient}$ is defined such that $\operatorname{stopgradient}(\bm{a}) = \bm{a}$ and $\nabla \operatorname{stopgradient}(\bm{a}) = 0$.
Combining the Gumbel-softmax trick with the $\operatorname{STargmax}$ results in the Straight-Through Gumbel Softmax (ST-GS) which gives discrete samples and with a usable gradient. The straight-through operator is biased but has low variance; in practice, it works very well and is better than the high-variance unbiased estimates one could get through REINFORCE~\citep{Havrylov2017}. In short, this trick allows us to train neural network models that incorporate fully discrete sampling operations using gradient-based methods in a fully end-to-end fashion.

To conclude this subsection we would like to highlight that autoencoders, variational autoencoders and many of the models used for exploring emergent communication with referential games are all inherently linked. All of these models attempt to compress raw data into a small number of latent variables, and thereby capture salient information, whilst discarding information which is not relevant to the task at hand. The only thing that is different in these models is the choice  of how the latent variables are modelled. In particular, the central part of the model by \citet{Havrylov2017} that we build upon in this paper (see Section~\ref{Section:ExperimentalSetup}), is essentially an autoencoder where the latent variable is a \textit{variable-length} sequence of categorical variables\footnote{the loss used is not one of reconstruction, however, it certainly strongly encourages the receiving agent to reconstruct the feature vector produced by the sender agent}; this is in many ways similar to the variational autoencoder variant demonstrated by \citet{DBLP:conf/iclr/JangGP17} which used \textit{fixed} length sequences of Bernoulli or categorical variables.

\subsection{Feature Learning}
% say something about traditional features vs convolutional feature maps
Among a variety of unsupervised approaches for feature representation learning, the self-supervised learning framework is one of the most successful as it uses pretext tasks such as image inpainting \citep{pathak2016context}, predicting image patches location \citep{doersch2015unsupervised} and image rotations \citep{gidaris2018unsupervised}. Such pretext tasks allow for the target objective to be computed without supervision and require high-level image understanding. As a result, high-level semantics are captured in the visual representations which are used to solve tasks such visual referential games. \citet{Kolesnikov_2019_CVPR} provide an extensive overview of self-supervised methods for feature learning. 
 
Recently, some of the most successful self-supervised algorithms for visual representation learning are using the idea of contrasting positive pairs against negative pairs. \citet{henaff2019data} tackles the task of representation learning with an unsupervised objective, Contrastive Predictive Coding \citep{oord2018CPC}, which extracts stable structure from still images. Similarly, \cite{ji2019} presents a clustering objective that maximises the mutual information between class assignments for pairs of images. They learn a neural network classifier from scratch which directly outputs semantic labels, rather than high dimensional representations that need external processing to be used for semantic clustering. Despite the recent surge of interest, \cite{chen2020simple} has shown through the strength of their approach that self-supervised learning still remains undervalued. They propose a simple framework, \textit{SimCLR},  for contrastive visual representation learning. SimCLR learns meaningful representations by maximising similarity between differently augmented views of the same image in the latent space. One of the main contributions of this work is that it outlines the critical role of data augmentations in defining effective tasks to learn useful representations. We will also explore this framework in some of our experiments detailed in Section~\ref{Subsec:Augumentation}.
% Unsupervised feature learning has been a long-studied problem in computer vision research \citep{gidaris2018unsupervised, henaff2019data, oord2018CPC, caron2018DeepCluster}. It is important to mention that in the setup we are interested in, the majority of the models used for extracting image features are based on convolutional neural networks (CNN). Among a variety of approaches, the self-supervised learning framework is one of the most successful as it uses pretext tasks such as image inpainting \citep{pathak2016context}, predicting image patches location \citep{doersch2015unsupervised} and image rotations \citep{gidaris2018unsupervised}. Such pretext tasks allow for the target objective to be computed without supervision and require high-level image understanding. As a result, high-level semantics are captured in the visual representations which are used to solve the tasks. \citet{Kolesnikov_2019_CVPR} provide an extensive overview of the area. 

Our attempt at encouraging the emergence of semantics in the learned communication protocol is most similar to previous works which combine multiple pretext tasks into a single self-supervision task \citep{chen2019self, Doersch_2017_ICCV}. Multi-task learning (MTL) rests on the hypothesis that people often apply knowledge learned from previous tasks to learn a new one. Similarly, when multiple tasks are learned in parallel using a shared representation, knowledge from one task can benefit the other tasks \citep{caruana1997multitask}. MTL has proved itself useful in language modelling for models such as BERT \citep{devlin2018bert} which obtains state-of-the-art results on eleven natural language processing tasks. More recently, \citet{radford2019language} combine MTL and language model pretraining, and propose MT-DNN, a model for learning representations across multiple natural language understanding tasks. In this work, we are also interested in the effect of solving multiple tasks on the semantics captured in the communication protocol.

% Multi-task (ex Language Models BERT)
% "Emergent language research rests on the hypothesis that, if we want
% deep networks to develop language-like communication skills, we cannot just
% train them to reproduce statistical regularities in static linguistic corpora (as is done in language modeling, e.g., Radford et al. [2019]), but we should plunge
% them into interactive scenarios, letting them develop a code to cooperatively
% solve their tasks[...]"
% Language models as unsupervised multi-task learners mentioned in Baroni's rat paper https://d4mucfpksywv.cloudfront.net/better-language-models/language-models.pdf

\section{Baseline Experimental Setup}\label{Section:ExperimentalSetup}
In this section we provide the details of our experimental setup; we start from \citet{Havrylov2017}'s image reference game. The objective of the game is for the Sender agent to communicate information about an image it has been given to allow the Receiver agent to correctly pick the image from a set containing many (127 in all experiments) distractor images.

\subsection{Model Architecture}\label{SubSec:modelarchi}
%attention, diff types of tasks
\begin{figure}
    \centering
    \resizebox{\textwidth}{!}{\input{model-orig.tikz}\unskip}
    \caption{\citet{Havrylov2017}'s game setup and model architecture.}
    \label{fig:emergentlanggame}
\end{figure}    

\citet{Havrylov2017}'s model and game are illustrated in Figure~\ref{fig:emergentlanggame}. The Sender agent utilises an LSTM to generate a sequence of tokens given a hidden state initialised with visual information and a Start of Sequence (SoS) token. To ensure that a sequence of only discrete tokens is transmitted, the output token logits produced by the LSTM cell at each timestep are sampled with the Straight-Through Gumbel Softmax operator (GS-ST).\footnote{\citet{Havrylov2017} experimented with ST-GS, the relaxed Gumbel Softmax and REINFORCE in their work, however, we focus our attention on ST-GS here.} 
%The ST-GS provides a one-hot vector at each time step but uses the gradients of the relaxed Gumbel Softmax during the backward pass to circumvent the problem that the sampling is non-differentiable~\citep{DBLP:conf/iclr/MaddisonMT17,DBLP:conf/iclr/JangGP17}. 
The Receiver agent uses an LSTM to decode the sequence of tokens produced by the Sender, from which the output is projected into a space that allows the Receiver's image vectors to be compared using the inner product. \citet{Havrylov2017} use a fixed VGG16 CNN pretrained on ImageNet to extract image features in both agents. The model is trained using a hinge-loss objective to maximise the score for the correct image whilst simultaneously forcing the distractor images to have low scores. The Sender can generate messages up to a given maximum length; shorter codes are generated by the use of an end of sequence (EoS) token. Although not mentioned in the original paper, we found that the insertion of a BatchNorm layer in the Sender between the CNN and LSTM, and after the LSTM in the Receiver, was critical for learnability of the model and reproduction of the original experimental results.   

\subsection{Training Details}\label{SubSec:TrainingDetails}
Our experiments use the model described above with some modifications under different experimental settings. In all cases, we perform experiments using the CIFAR-10 dataset rather than the COCO dataset used in the original work (to replicate the original results requires multiple GPUs due to the memory needed, as well as considerable training time\footnote{We found that about 32GB of RAM spread across four RTX-2080Ti GPUs was required with the sender, receiver and feature extractor each being placed on a different GPU, and the loss being computed on the forth. Each epoch of 74624 games (for each batch of 128 images we played the 128 possible games by taking each image in turn as the target) took around 7 minutes to complete. The convergence of the communication rate to a steady level took at least 70 epochs.}). In light of the smaller resolution images and lower diversity of class information, we choose a word embedding dimension of 64, hidden state dimension of 128, and total vocabulary size of 100 (including the EoS token). We also limit the maximum message length to 5 tokens. The training data is augmented using color jitter ($p_{bri}=0.1, p_{con}=0.1, p_{sat}=0.1, p_{hue}=0.1$), random grayscale transformation ($p=0.1$), and random horizontal flipping ($p=0.5$), so there is very low probability of the model seeing exactly the same image more than once during training. The batch size is set to 128, allowing for the Receiver to see features from the target image plus 127 distractors. Most simulations converge or only slowly improve after about 60 epochs, however for consistency, all results are reported on models trained to 200 epochs where convergence was observed to be guaranteed for well-initialised models\footnote{Certain model configurations were more sensitive to initialisation; this is discussed further in the next section.}. 

\subsection{Metrics}\label{Subsec:metrics}
Our key objective is to measure how much visual semantic information is being captured by the emergent language. If humans were to play this game, it is clear, as discussed in Section~\ref{Subsec:humanSemantics}, that a sensible strategy would be to describe the target image by its semantic content (e.g. ``a yellow car front-on'' in the case of the example in Figure~\ref{fig:emergentlanggame}). It is also reasonable to assume in the absence of strong knowledge about the make-up of the dataset (for example, that the colour yellow is relatively rare) that a semantic description of the object in the image (a ``car'') should have a strong part to play in the communicated message if visual semantics are captured. Work such as \citet{hare2006mind} considers the semantic gap between object/class labels and the full semantics, significance of the image. However, in the case of the CIFAR-10 dataset in which most images have a single subject, ``objectness'' can be considered a reasonable measure of semantics. 

With this in mind, we can measure to what extent the communicated messages capture the object by looking at how the target class places in the ranked list of images produced by the Receiver. More specifically, in the top-5 ranked images guessed by the Receiver, we can calculate the number of times the target object category appears, and across all the images we can compute the average of the ranks of the images with the matching category. In the former case, if the model captures more semantic information, the number will increase; in the latter, the mean-rank decreases if the model captures more semantic information. A model which is successful at communicating and performs almost ideal hashing would have an expected top-5 number of the target class approaching 1.0 and expected average rank of 60, whilst a model that completely captures the ``objectness'' (and still guesses the correct image) would have an expected top-5 target class count of 5 and expected mean rank of 7.35. In addition to these metrics for measuring visual semantics, we also measure top-1 and top-5 communication success rate (receiver guesses correctly in the top-1 and top-5 positions) and the message length for each trial. On average across all games, there are 13.7 images with the correct object category in each game (on the basis that the images are uniformly drawn without replacement from across the 10 classes and the correct image and its class are drawn from within this). If the message transmitted only contained information about the object class, then the communication success, when considering the top-1 and top-5 choices of the Receiver, would be on average 0.07, and 0.36 respectively. Since we observe that throughout the experiments there is a significant trade-off between the semantics measures and the top-1 communication rate, we consider top-5 rate a better indication of the capacity of the model to succeed at the task while learning notions of semantics. If the communication rate in top-5 is higher than the average, it means that the message must contain additional information about the correct image, beyond the type of object. However, we do not easily have the tools to find out what that extra information might be; it could be visual semantics such as attributes of the object, but it could also be some robust hashing scheme.

\section{Experiments, Results and Discussion}
\label{Section:Experiments}

This section describes a number of experiments and investigations into the factors that influence the emergence of visual semantics in the baseline experimental setup described in the previous section, as well as extended versions of that baseline model. We start by exploring to what extent using a pretrained feature extractor influences what the agents learn to communicate and then look at different ways in which semantically meaningful communication can be encouraged without any form of supervision (including supervised pretraining).

\subsection{The effect of different weights in the visual feature extractor}\label{Subsec:diffWeight}
Generating and communicating hash codes is very clearly an optimal (if very unhuman) way to play the image guessing game successfully. In ~\citet{Havrylov2017}'s original work there was qualitative evidence that this did not happen when the model was trained, and that visual semantics were captured. An important first question is: to what extent is this caused by the pretrained feature extractor?

\begin{figure}
    \centering
    % PLOTS OF LEARNING OVER EPOCHS OF BASIC MODELS [Pre/Rnd/E2E] SHOWING VARIANCE of [loss/T1CR/T5CR/TCT5/TCAR]
    \includegraphics[width=\textwidth]{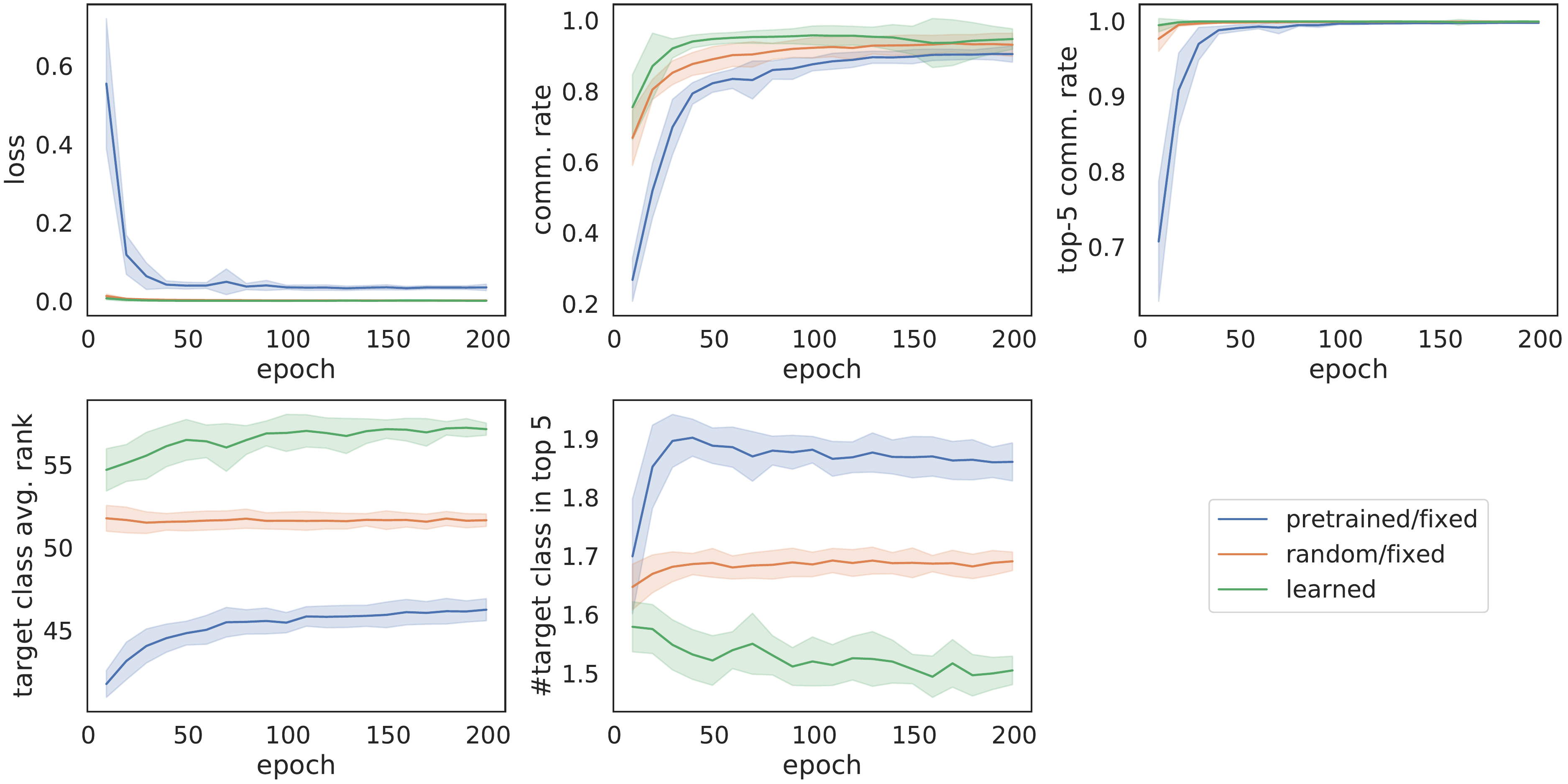}
    \caption{The game-play and semantic performance over the training epochs of the three model variants using a: pretrained, random or fully learned feature extractor CNN. The loss plot shows that the learned and random models converge much faster than the pretrained one, and have lower variance allowing the agents to evolve a successful game strategy.}
    \label{fig:learning}
\end{figure}

We attempt to answer this question by exploring three different model variants: the original model with the CNN fixed and initialised with ImageNet weights; the CNN fixed, but initialised randomly; and, the CNN initialised randomly, but allowed to update its weights during training. 
Results from these experiments are summarised in Table~\ref{table:weights}. The first observation relates to the visual-semantics measures; it is clear (and unsurprising) that the pretrained model captures the most semantics of all the models. It is also reasonable that we observe less semantic alignment with the end-to-end model; without external biases, this model should be expected to move towards a hashing solution. It is perhaps somewhat surprising however that the end-to-end model and the random model have a similar communication success rate, however, it is already known that a randomly initialised CNN can provide reasonable features~\citep{Saxe:2011:RWU:3104482.3104619}. During training, the Sender and Receiver convergence had particularly low variance with both the end-to-end and random models, allowing the agents to quickly evolve a successful strategy. This is in contrast to the pretrained model which had markedly higher variance as can be seen from the plots in Figure~\ref{fig:learning}.

One might question if the end-to-end model would be handicapped because it had more weights to learn in the same number of epochs (200 for all models), however, as the results show, the end-to-end model has the best performance. We also investigated if the models required more training time; nevertheless, training all the models for 1000 epochs yielded only a $2\%$ improvement in communication rate across the board.

% The values are reported after 200 epochs of training when the model has converged.

\begin{table}
  \caption{The effect of different weights in the feature extractor CNN. Measures are averaged across 7 runs of the game for each model on the CIFAR-10 validation set. Communication rate values in brackets are standard deviations across games,  which show the sensitivity to different model initialisations and training runs. The message length standard deviation is measured across each game and averaged across the 7 runs, and show how much variance there is in transmitted message length.}
  \label{table:weights}
  \centering
   \begin{tabular}{llllll}
    \toprule
     Feature extractor & Comm. & Message & Top-5 & \#Target  & Target \\
       & rate & length & comm. & class  & class \\
       & & & rate & in top-5 & avg. rank \\ 
    \midrule
     Pretrained \& fixed &0.90 ($\pm$0.02) & 4.93 ($\pm$0.34) & 1 & 1.86 & 46.25\\
 
     Random \& frozen & 0.93 ($\pm$0.03) &4.90 ($\pm$0.39) & 1 & 1.69 &51.65\\
 
    Learned end-end & 0.94 ($\pm$0.02) &4.90 ($\pm$0.39) & 1 & 1.5 &57.14\\
    \bottomrule
  \end{tabular}
\end{table}

\begin{table}
  \caption{The effect of different weights in the feature extractor CNN when the model is augmented by adding noise and/or random rotations to the Sender agent's input images, and when independently augmenting both agent's inputs images following the SimCLR framework \citep{chen2020simple}. Measures as per Table~\ref{table:weights}.}
  \label{table:noiserot}
  \centering
   \begin{tabular}{llllll}
    \toprule
     Feature extractor & Comm. & Message & Top-5 & \#Target  & Target
     \\
       & rate & length & comm. & class & class \\
       & & & rate & in top-5 & avg. rank \\
    \midrule
    \multicolumn{6}{l}{\textbf{Sender images augmented with Gaussian noise:}}\\
      Pretrained \& fixed & 0.89 ($\pm$0.02) & 4.93 ($\pm$0.33) & 0.99 & 1.86 & 46.39\\

      Random \& frozen & 0.94 ($\pm$0.01) & 4.90 ($\pm$0.38) & 1 & 1.66 & 52.45\\

      Learned end-end & 0.94 ($\pm$0.02) & 4.92 ($\pm$0.33) & 1 & 1.51 & 57.33\\
    \midrule
    \multicolumn{6}{l}{\textbf{Sender images augmented with random rotations:}}\\
      Pretrained \& fixed & 0.8 ($\pm$0.05) & 4.94 ($\pm$0.32) & 0.99 & 2.03 & 42.9\\

      Random \& frozen & 0.80 ($\pm$0.12) & 4.87 ($\pm$0.45) & 0.98 & 1.7 & 51.43\\

      Learned end-end & 0.92 ($\pm$0.04) & 4.92 ($\pm$0.32) & 1 & 1.59 & 55.8\\
     \midrule
     \multicolumn{6}{l}{\textbf{Sender images augmented with Gaussian noise and random rotations:}}\\
      Pretrained \& fixed & 0.76 ($\pm$0.02) & 4.92 ($\pm$0.38) & 0.98 & 2.01 & 42.85\\

      Random \& frozen & 0.67 ($\pm$0.26) & 4.77 ($\pm$0.57) & 0.92 & 1.62 &51.37\\

      Learned end-end & 0.90 ($\pm$0.06) & 4.94 ($\pm$0.29) & 1 & 1.58 & 55.8\\
     \midrule
     \midrule
     \multicolumn{6}{l}{\textbf{Sender \& Receiver images independently augmented (SimCLR-like):}}\\
      Pretrained \& fixed & 0.48 ($\pm$0.03) & 4.90 ($\pm$0.41) &0.86 & 2.14 & 38.08\\
 
     Random \& fixed & 0.42 ($\pm$0.10) & 4.92 ($\pm$0.33) & 0.85 & 1.68 & 47.94\\
 
     Learned end-end & 0.72 ($\pm$0.05) & 4.91 ($\pm$0.39) & 0.98 & 2.00 & 42.37\\
    \bottomrule
  \end{tabular}
\end{table}

\subsection{Making the game harder with augmentation}\label{Subsec:Augumentation}
We next investigate the behaviour of the same three model variants while playing a slightly more difficult game. The input image to the Sender is randomly transformed, and thus will not be pixel-identical with any of those seen by the Receiver. For the model to communicate well it must either capture the semantics or learn to generate highly-robust hash codes.

\paragraph{Noise and Rotation.}
We start by utilising transformations made from random noise and random rotations. The added noise is generated from a normal distribution with mean 0 and variance 0.1, and the rotations applied to the input images are randomly chosen from \{\ang{0}, \ang{90}, \ang{180}, \ang{270}\}. 

The first part of Table~\ref{table:noiserot} shows the effect of adding either noise or rotations, or both. In general noise results in a slight increase in the communication success rate. More interestingly, for randomly rotated Sender images the augmentation tends to increase the visual semantics captured by all the models, although this is most noticeable in the pretrained variant. At the same time, the communication success rate of the pretrained model drops; it is an open question as to whether this could be resolved by sending a longer message. Finally, the models augmented with both noise and rotations do no show any improvement over the rotation only game in terms of the semantics measure. As one might guess, noise only makes the game harder, a fact which is reflected in the slight drop of communication success, but does not explicitly encourage semantics.

\paragraph{More complex transformations.}
We continue by adding a more complex composition of data augmentations to the game. \citet{chen2020simple} have recently shown that combinations of multiple data augmentation operations have a critical role in contrastive self-supervised learning algorithms and improve the quality of the learned representations. We implement their transformation setup in our game, with sender and receiver having differently augmented views of the same image. We follow the combination proposed by \citeauthor{chen2020simple} for the CIFAR-10 experiment which consists in sequentially applying: random cropping (with flip and resize to the original image size) and random colour distortions\footnote{The details of the data augmentations are provided in the appendix of \citet{chen2020simple} and available at \url{https://github.com/google-research/simclr}}. We test if the combination does improve the learned representations in a self-supervised framework as ours, which however does not use a contrastive loss in the latent space, but the aforementioned hinge-loss objective (see Section~\ref{SubSec:modelarchi}). It is also worth noting that we continue using a VGG-16 feature extractor, as opposed to the ResNet \citep{he2016deep} variants used by \citet{chen2020simple}. The game is played as described in Section~\ref{Section:ExperimentalSetup}, but this time each image is randomly transformed twice, giving two completely independent views of the same example, and hence, making the game objective harder than with the noise and rotation transformations\footnote{In the noise and rotation case only the sender's image was transformed. It is conceivable in this case that the sender might learn to de-noise or un-rotate the feature in order to establish a communication protocol. If images are transformed on both sides of the model, the agents won't have an easy way of learning a `correct' inverse transform.}. 

The lower part of Table~\ref{table:noiserot} shows the results of the newly-augmented game for the different configurations of feature extractors used previously (pretrained with ImageNet and fixed; random and fixed; and, learned end-to-end). The results show that, indeed, by extending the augmentations and composing them randomly and independently for Sender and Receiver, the communication task becomes harder, hence the communication success is lower than in the previous experiments. However, as \citet{chen2020simple}'s results have also shown, the quality of the representations improves considerably, especially for the model `Learned end-end', and this is reflected in the improvement of our measures for the amount of semantic information captured in the learned communication protocol. Specifically, the number of times the target class appears in top-5 predictions increases by almost half a point for the pretrained and learned model, and the average rank of the target class lowers (over 10 units for the learned model) which indicates that the protocol captures more content information and is less susceptible to only hashing the images. Using this approach, the learned model achieves the highest communication success while also getting semantic results close to the model with an ImageNet pretrained feature extractor.

% In addition, we tested playing the game with a feature extractor pretrained in the completely self-supervised framework of \citep{chen2020simple}, while allowing the weights to modify or fixing them during the game. Overall, the results show that the Top-1 communication success is lower for the models which use or start with a pretrained feature extractor (either the ImageNet classification task or the SimCLR self-supervised objective). However, their top-5 communication rate and their semantics metrics, which overrun all other tested configurations, suggest that these model capture more content related information and diverge from a hashing approach while still managing to solve the game task.

% The simplicity of the framework proposed by \citet{chen2020simple} allows us to transfer it to a different model architecture and use it with a different loss, but somewhat similar in what it aims to achieve. Using their combination of augmentations is to some extent similar to imposing an additional task which we explore next.
It is particularly interesting to observe that by the relative simplicity of applying the same transformations to the images as \citet{chen2020simple} we encourage semantic alignment in a completely different model architecture and loss function. This suggests that the value of \citet{chen2020simple}'s proposal for contrastive learning is more towards the choice of features rather than the particular contrastive loss methodology.

\subsection{Making the game harder with multiple objectives}\label{Subsec:multiple-objec}

\begin{figure}
    \centering
    \resizebox{0.7\textwidth}{!}{\input{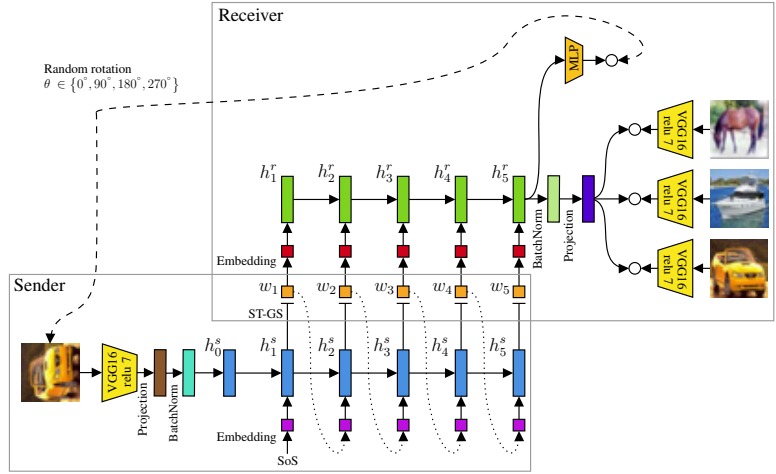}\unskip}
    \caption{Extended game with the Receiver also required to guess the orientation of the Sender's image.}
    \label{fig:emergentlanggame-rot}
\end{figure}

\begin{figure}[h]
    \centering 
    \resizebox{0.7\textwidth}{!}{\input{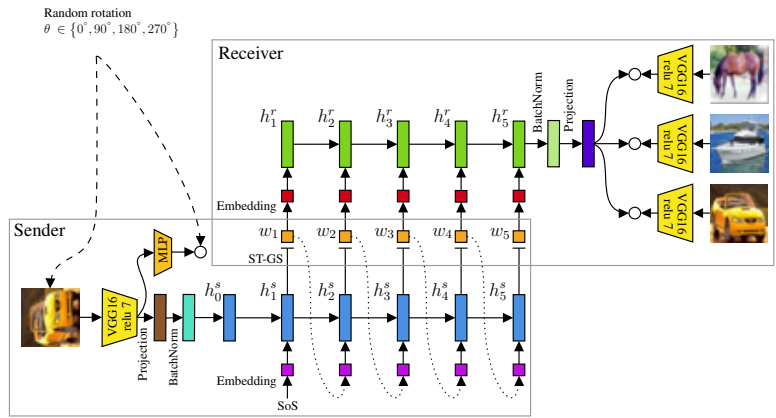}\unskip}
    \caption{Extended game with the Sender augmented with an additional loss based on predicting the orientation of the input image.}
    \label{fig:emergentlanggame-srot}
\end{figure}

The experimental results with the model setups shown in Tables~\ref{table:weights} and \ref{table:noiserot} clearly show that the fully-learned models always collapse towards game-play solutions which are not aligned with human notations of visual semantics. Conversely, the use of a network that was pretrained in a supervised fashion to classify real-world images has a positive effect on the ability of the communication system to capture visual semantics. On the other hand, using a different experimental setup involving a complex set of independent transformations of the images given to the sender and receiver helps the learned model acquire and use more of the visual-semantic information, similar to the pretrained model. However, this improvement comes at the cost of reducing the communication success rate as the game becomes much harder when using the proposed augmentations. 

We continue by exploring if it might be possible for a communication protocol with notions of visual semantics to emerge directly from pure self-supervised game-play. In order to achieve this, we propose that the agents should not only learn to play the referential game, but they should also be able to play other games (or solve other tasks). In our initial experiments we formulate a setup where the agents not only have to play the augmented version of the game described in Section~\ref{Subsec:Augumentation} (with both noise and rotations randomly applied to the image given to the Sender, but not the Receiver), but also one of the agents has to guess the rotation of the image given to the Sender as shown in Figures~\ref{fig:emergentlanggame-rot} and~\ref{fig:emergentlanggame-srot}. 

This choice of the additional task is motivated by \citet{gidaris2018unsupervised} who showed that a self-supervised rotation prediction task could lead to good features for transfer learning, on the premise that in order to predict rotation the model needed to recognise the object. The rotation prediction network consists of three linear layers with Batch Normalisation before the activation functions. The first two layers use ReLU activations, and the final layer uses a Softmax to predict the probability of the four possible rotation classes. Except for the final layer, each layer outputs 200-dimensional vectors. Cross-Entropy is used as the loss function for the rotation prediction task ($\mathcal{L}_{rotation}$). All other model parameters and the game-loss definition match those described in Section~\ref{Section:ExperimentalSetup}.

\begin{table}
  \caption{End-to-End learned models with an additional rotation prediction task. Measures as per Table~\ref{table:weights}, except for the inclusion of the accuracy of rotation prediction.}
  \label{table:extmodels}
  \centering
   \begin{tabular}{llllll}
    \toprule
     Model & Comm. & Top-5 & \#Target   & Target & Rot.\\
       & rate & comm.  & class & class & acc.\\
       & & rate & in top-5 & avg. rank & \\
    \midrule
      Receiver-Predicts (Fig.~\ref{fig:emergentlanggame-rot}) & 0.58 & 0.96 & 1.85 & 48.75 & 0.80\\
      Sender-Predicts (Fig.~\ref{fig:emergentlanggame-srot}) & 0.72 & 0.98 & 2.05 & 42.89 & 0.83\\
    \bottomrule
  \end{tabular}
\end{table}

Results of these experiments are shown in Table~\ref{table:extmodels}. We ran a series of experiments to find optimal weightings for the two losses such that the models succeed at the communication task while also acquiring notions of visual semantics. Both experiments presented, with the Sender-predicts model (Figure~\ref{fig:emergentlanggame-srot}) and the Receiver-predicts model (Figure~\ref{fig:emergentlanggame-rot}), used a weighted addition $0.5\cdot\mathcal{L}_{rotation} + \mathcal{L}_{game}$, where $\mathcal{L}_{game}$ refers to the original hinge-loss objective for the game proposed by \citet{Havrylov2017}. For the latter model we also tried using additive loss with learned weights (following~\citet{kendall2017multi}) however this created a model with good game-play performance, but an inability to predict rotation (and poor semantic representation ability). 

Training these models is harder than the original Sender-Receiver model because the gradients pull the visual feature extractor in different directions; the game achieves good performance when the features behave like hash codes, whereas the rotation prediction task requires much more structured features. This conflict means that it is difficult to train the models such that they have the ability to solve both tasks. Clearly further work in developing optimisation strategies for these multi-game models is of critical importance in future work.

Whilst there is still a way to go to achieve the best levels of game-play performance shown in Tables~\ref{table:weights} and~\ref{table:noiserot}, it is clear that these fully self-supervised end-to-end trained models can both learn a communication system to play the game(s) that diverges from a hashing solution towards something that better captures semantics. The lower game-play performance might however just be a trade-off one has to live with when encouraging semantics with a fixed maximum message length; this is discussed further at the end of the following subsection.

\subsection{Playing games with self-supervised pretraining}\label{Sec:SS-pretraining}

Having observed that a completely learned model, with the right augmentations or instructed to solve multiple tasks which enforce notions of `objectness', can already acquire some visual semantics, we end by exploring the effect of combining these two approaches: the multi-task game described in Section~\ref{Subsec:multiple-objec} with the previously mentioned self-supervised SimCLR framework \citep{chen2020simple}. The goal of this is to test whether a pretrained feature extractor, also trained on a task which does not require human intervention, can further improve the meaning of the communication protocol, pushing it towards a more human-like version. This set of experiments was performed with the Sender-Predicts model described in Section~\ref{Subsec:multiple-objec}. We employ independent augmentations for the Sender and Receiver agents that match those detailed in the second half of Section \ref{Subsec:Augumentation}. To some extent, this resembles \citeauthor{Lowe*2020On}'s Supervised Self-Play approach in which self-play in a multi-agent communication game and expert knowledge are interleaved. In our case, however, the VGG16 feature extractor network was pretrained with \citet{chen2020simple}'s framework in a completely self-supervised way. 

\begin{table}
  \caption{The effect of interleaving self-supervision and multi-agent game-play. The game setup has two tasks: Sender Predicting Rotation as per Table~\ref{table:extmodels}, while using various augmentations (original and SimCLR same or individual).}
  \label{table:VGGSenderPredsimCLR}
  \centering
   \begin{tabular}{llllll}
    \toprule
     Feature Extractor & Comm. & Top-5 & \#Target   & Target & Rot.\\
       & rate & comm.  & class & class & acc.\\
       & & rate & in top-5 & avg. rank & \\
    \midrule
    \multicolumn{6}{l}{\textbf{Sender \& Receiver images augmented with the original transforms:}}\\
      Learned end-end & 0.72  & 0.98 & 2.05 & 42.89 & 0.83\\
      Pretrained SS end-end & 0.84 & 0.99 & 2.19 & 40.19 & 0.79\\
      Pretrained SS \& fixed & 0.80 & 0.99 & 2.23 & 39.72 & 0.7\\
    \midrule
    % \multicolumn{6}{l}{\textbf{Sender \& Receiver images augmented with the same SimCLR transforms:}}\\
    %   Learned end-end & 0.40 ($\pm$0.49) & 0.81 & 1.86 & 41.44 & 0.82\\
    %   Pretrained SS end-end & 0.68 ($\pm$0.46) & 0.96 & 2.06 & 41.68 & 0.79\\
    % \midrule
    \multicolumn{6}{l}{\textbf{Sender \& Receiver images augmented with SimCLR transforms:}}\\
      Learned end-end & 0.53  & 0.92 & 2.22 & 37.16 & 0.80\\% this hasnowd
    %   Learned(SimCLR-individ-wd) & 0.54 ($\pm$0.49) & 0.92 & 2.21 & 37.35 & 0.81\\
      Pretrained SS end-end & 0.49 & 0.89 & 2.18 & 38.74 & 0.79\\
      Pretrained SS \& fixed & 0.42  & 0.85 & 2.14 & 39.57 & 0.78\\
    \bottomrule
  \end{tabular}
\end{table}

The results of the multi-objective game played with the Sender-predicts model, in the initial setup and with the modified SimCLR transforms, are presented in Table~\ref{table:VGGSenderPredsimCLR}. We compare the different type of weights in the feature extractor again: learned end-to-end, pretrained in a self-supervised way and fixed, or allowed to change during the game-play. For the games which only start with a self-supervised pretrained VGG16, we chose to fix the weights of the feature extractor for the first 5 epochs before allowing any updates. This was based on empirical results which showed that it helped to stabilise the LSTM and Gumbal-softmax part of the models before allowing the gradients to flow through the pretrained feature extractor part. We hypothesise that this is due to the risk of bad initialisation in the LSTMs which can cause the models to fail to converge at the communication task. This observation can be generalised over all the experiments in this work, as all the models with a fixed feature extractor appear to be slightly more unstable than those with learned ones, in contrast to fully learned models which always converged (see Figure~\ref{fig:learning}).

As the results show, the model which best captures visual semantics is the one learned end-to-end using the SimCLR transforms. It is again obvious that between the two setups, the second makes the game significantly harder as the agents are now also required to extract and encode information about the object orientation, on top of seeing independently augmented input images. This is reflected in the drop of the top-1 communication success, although this does not hold for the top-5 rate. If the semantics improve, it implicitly means that more of the object category is captured in the learned language which diverges from a hashing protocol. As previously mentioned in Section~\ref{Subsec:metrics}, if the model only transmitted information about the object, the top-5 communication rate would be on average 0.36. Since this metric is significantly higher, it implies that the message must contain additional information, beyond the type of object. This could be visual semantics such as attributes of the object, but it could also just be a more robust hashing scheme based on pixel or low-level feature values. 

Another interesting observation is that using a self-supervised pretrained feature extractor, in the original setup, helps improve communication success and the semantics measures at the same time. This finding confirms that self-supervised pretraining in this type of game can be as beneficial, or even better, as the supervised pretraining on ImageNet used in a less complex variant of the game (see Table~\ref{table:noiserot}). %One direction for future work should be examining in more depth how self-supervision and self-play updates can be combined in this sort of environment. 

% \section{Questions to answer}

% \begin{itemize}

%     \item Which structures promote the emergence of language, as opposed to protocols that are not inherently language like (e.g. hash maps), both in humans as well as in artificial neural nets? 
%     \item What happens when we don't ground the visual system, can we still understand what's going on?
%     \item Effect of pretrained visual systems vs learning it
%     \item Is this type of loss encouraging emergent communication?
% \end{itemize}

\section{Conclusions and Future Work}
\label{Section:Concl}
In this paper, we have explored different factors that influence the human interpretability of a communication protocol, that emerges from a pair of agents learning to play a referential signalling game with natural images. We first quantify the effect that using a pretrained visual feature extractor has on the ability of the language to capture visual semantics. We empirically show that using pretrained feature extractor weights from a supervised task inductively biases the emergent communication channel to become more semantically aligned, whilst both random-fixed and learned feature extractors have less semantic alignment, but better game-play ability due to their ability to learn hashing schemes that robustly  identify particular images using very low-level information.

We then perform an analysis of the effect that different forms of data augmentation and transformation have on the agents' ability to communicate object type related information. Inducement of zero-mean Gaussian noise into the sender's image does not serve to improve the semantic alignment of messages but does perhaps have a mild effect of improving the robustness of the hashing scheme learned by the models. The addition of rotation to the sender's image results in a mild improvement in the semantic alignment, although in the case of the models with fixed feature extractors this is at the cost of game-play success rate. More complex combinations of data transforms applied independently to the sender's and receiver's images, are demonstrated to give a sizeable boost to the visual semantic alignment for the model learned in an end-to-end fashion. 

We then demonstrate that it is possible to formulate a multiple-game setting in which the emergent language is \textit{more} semantically grounded also without the need for any outside supervision. We note these models represent difficult multi-task learning problems, and that the next steps in this direction would benefit from full consideration of multi-task learning approaches which deal with multiple objectives that conflict~\citep[e.g.][]{SenerNips2018, kendall2017multi}. 

Finally, we have shown that pretraining the visual feature extractor on a self-supervised task, such as that of \citet{chen2020simple}, can further improve the quality of the semantics notions captured by a fully learned model. One way of looking at self-supervised pretraining is to consider it as self-play of a different game, before engaging in the main communication task/game. From this point of view, further work in the area of emergent communication should explore other combinations of self-supervised tasks. Creating environments in which agents have to solve multiple tasks, concurrently or sequentially, while using the correct type of data augmentations seems to balance the trade-off between performing the task well and developing a communication protocol interpretable by humans. As \citet{Lowe*2020On} has also shown, interleaving supervision and self-play can benefit multi-agent tasks while reducing the amount of necessary human intervention. 

Clearly there are many research directions that lead on from the points we have highlighted above. We, however, would draw attention to perhaps the two most important ones: better disentanglement and measurement of semantics; and more investigations into the role of self-play with multiple tasks.

If emergent communication channels are to be truly equatable to the way that humans communicate performing similar tasks, then we need to build models that more clearly disentangle different aspects of the semantics of the visual scenes they describe. Although throughout the paper we have used `objectness' as an initial measure of semanticity, we would be the first to admit how crude this is.
We have highlighted in the discussion of results, that when a model has both high semantics (using our objectness measures) and high game-play success rates we do not know what kind of information is being conveyed, in addition to information about the object, to  allow the model to successfully play the game; it could be information about semantically meaningful object attributes (or even other objects in the scene), or it could just be some form of robust hash code describing the pixels. The reality of current models is that it's probably somewhere in between, but it is clear that what is needed is a better-formalised strategy to distinguish between the two possibilities. We suspect that to achieve this we require a much more nuanced dataset with very fine-grained labels of objects and their attributes. This would then ultimately allow the challenge of disentangling meaningful semantic attribute values in the communication protocol to be addressed.

Our experimental results clearly show that pretraining, which can be seen as a form of self-play, can clearly benefit a model. Building upon these results we would like to encourage further research in the emergent communication area to consider self-supervision as additional games which can be combined with the communication task as a way of encouraging human-interpretability of emergent communication protocols. Such a direction seems entirely natural given what is known and has been observed about how human infants learn.
\bibliographystyle{apacite}
\bibliography{refs.bib}

\end{document}